\pdfoutput=1

\documentclass[11pt]{article}

\usepackage[]{acl}

\usepackage{hyperref}
\usepackage{url}

\usepackage{times}
\usepackage{latexsym}

\usepackage[T1]{fontenc}

\usepackage[utf8]{inputenc}
\usepackage{microtype}

\usepackage{booktabs}
\usepackage{subcaption}
\usepackage{url}
\usepackage{graphicx}
\usepackage[normalem]{ulem}
\usepackage{algorithm}
\usepackage{algorithmic}
\usepackage{amsmath,amssymb}
\usepackage{longtable}
\usepackage{lscape}

\usepackage{lineno}

\usepackage{comment}
\usepackage{todonotes}
\usepackage{color,soul}

\title{QCRI's COVID-19 Disinformation Detector:\\ A System to Fight the COVID-19 Infodemic in Social Media}

\author{Preslav Nakov$^1$, Firoj Alam$^1$, Yifan Zhang$^1$, Animesh Prakash$^2$, Fahim Dalvi$^1$\\
  $^1$Qatar Computing Research Institute, HBKU, Doha, Qatar \\
  $^2$Vellore Institute of Technology\\
  \texttt{\{pnakov, falam, yzhang, faimaduddin\}@hbku.edu.qa},\\ \texttt{Animesh\_Prakash@hotmail.com} 
  \\}

\begin{document}
\maketitle
\begin{abstract}
Fighting the ongoing COVID-19 infodemic has been declared as one of the most important focus areas by the World Health Organization since the onset of the COVID-19 pandemic. While the information that is consumed and disseminated consists of promoting fake cures, rumors, and conspiracy theories to spreading xenophobia and panic, at the same time there is information (e.g., containing advice, promoting cure) that can help different stakeholders such as policy-makers. Social media platforms enable the infodemic and there has been an effort to curate the content on such platforms, analyze and debunk them. While a majority of the research efforts consider one or two aspects (e.g., detecting factuality) of such information, in this study we focus on a multifaceted approach, including an API,
\footnote{\url{https://app.swaggerhub.com/apis/yifan2019/Tanbih/0.8.0}} and a demo system,\footnote{\url{https://covid19.tanbih.org/}} which we made freely and publicly available. We believe that this will facilitate researchers and different stakeholders. A screencast of the API services and demo is available at \url{https://youtu.be/zhbcSvxEKMk}.
\end{abstract}

\section{Introduction}
\label{sec:introduction}

\begin{figure}[!htb]
	\centering
	\includegraphics[width=0.4\textwidth]{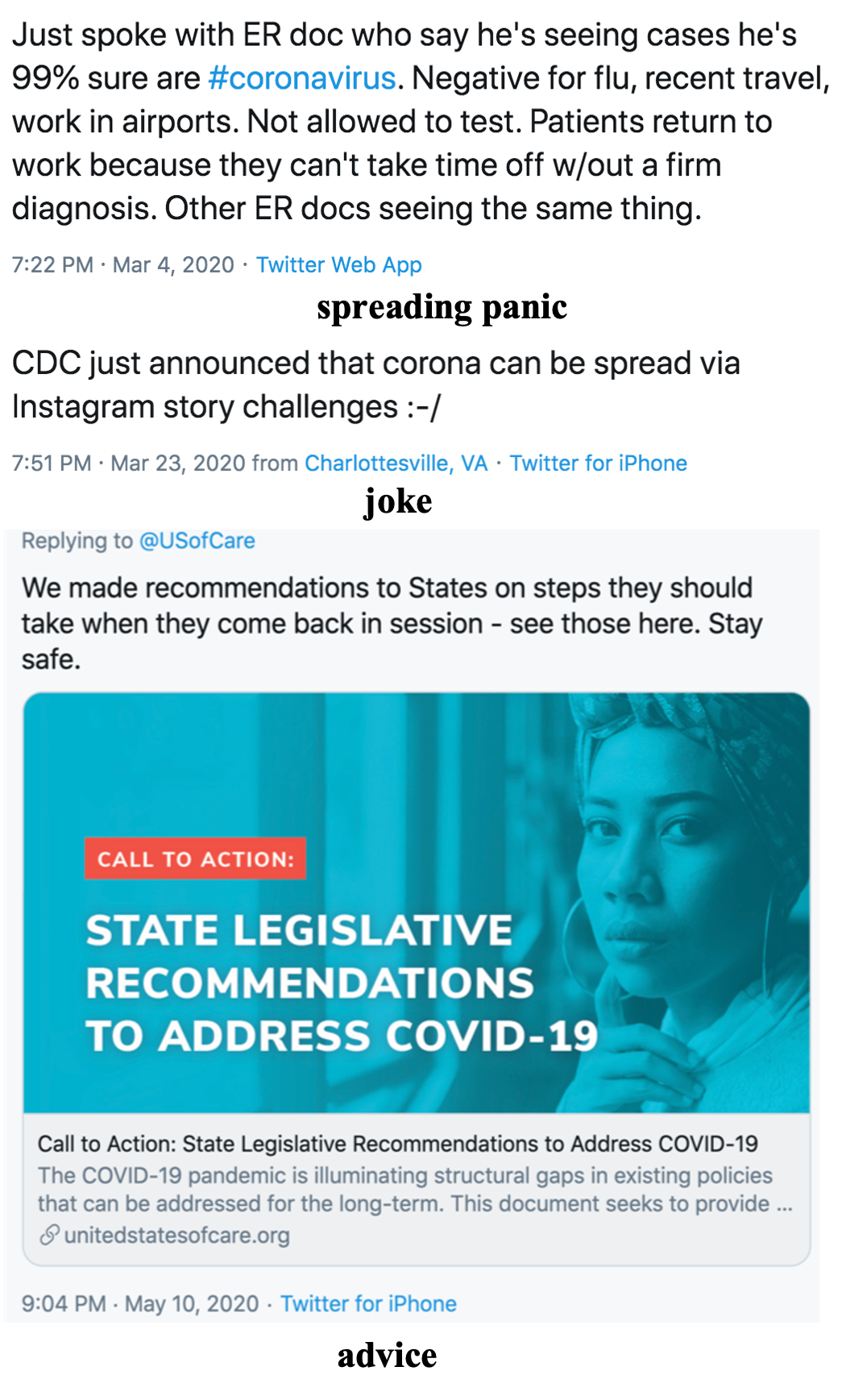}
	\caption{Examples of tweets that contain \textit{panic}, \textit{joke}, and \textit{advice}.}
	\label{fig:tweet_example}
\end{figure}

Since the emergence of the COVID-19 pandemic, there has been a rise in the spread of medical and political disinformation, which resulted in the \emph{first global infodemic}. Such information goes viral on social media and has been misleading to a large population. 
Alongside, there has also been information (e.g., containing advice, discussing action taken, and suggesting actions) that could be helpful for different stakeholders. The interplay between such diverse sets of information is unprecedented and requires a holistic approach that can facilitate journalists, fact-checkers, policymakers, social media platforms, and society. Figure~\ref{fig:tweet_example} shows some examples of the tweets that highlights how social media users discuss topics related to COVID-19 such as spreading panic, joke, and contains advice that can be useful for policy-makers. Addressing such multi-faceted aspects requires following a holistic approach. There have been efforts to address such multi-faceted aspects in terms of developing annotation guidelines and datasets in multiple languages \cite{alam2020fighting,song2021classification}.  

\begin{figure*}[!htb]
	\centering
	\includegraphics[width=0.85\textwidth]{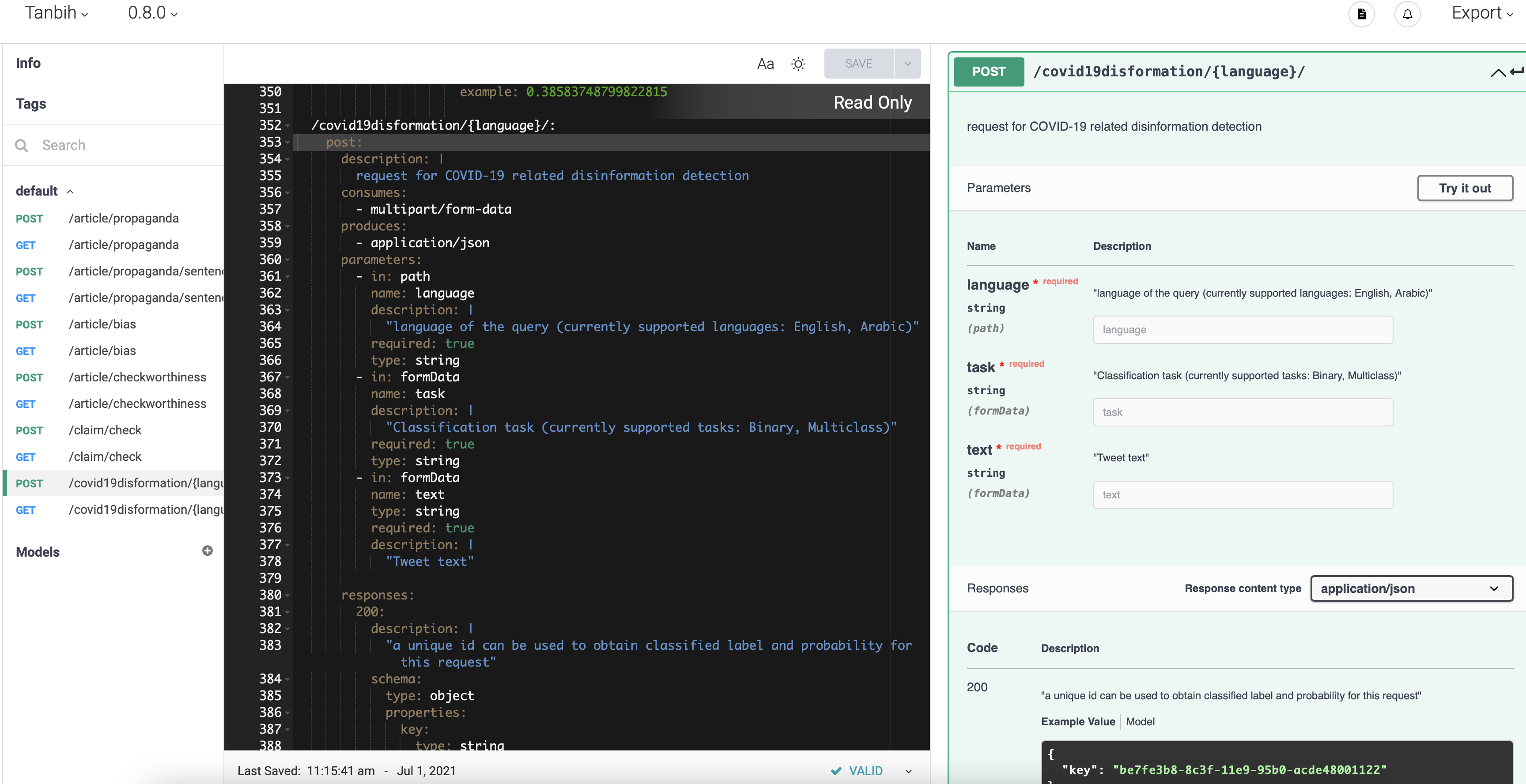}
	\caption{API services to process COVID-19-related tweets.}
	\label{fig:system_api}
\end{figure*}

\begin{figure*}[!tbh]
	\centering
	\includegraphics[width=1.0\linewidth]{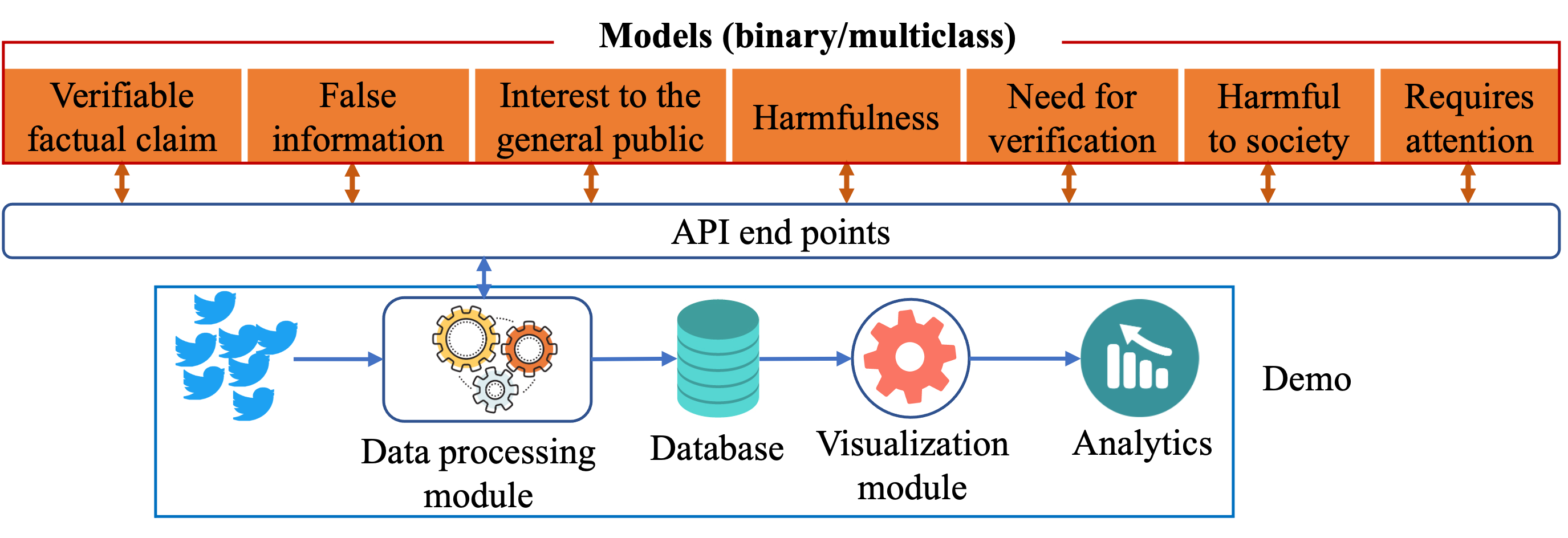}
	\caption{COVID-19 infodemic API services and data processing pipeline for the demo. The arrows indicate the information flow.}
	\label{fig:api_services}
\end{figure*}

While there has been significant research in developing resources (specifically creating labeled and unlabeled datasets), there has been a limited effort in developing publicly available systems, which are could be useful for general users as well for non-technical domain experts. To address this gap, in this paper, we describe COVID-19 Infodemic system -- API services (Figure \ref{fig:api_services}), and a demo (Figure \ref{fig:system_demo}), which can assist both technical and non-technical end users. Such services will be helpful to understand whether the information is correct, harmful, calling for action to be taken by relevant authorities, etc. Our motivation to develop such API services also came from the interest of policymakers, such as the Ministry of Public Health of a country. Note that our developed APIs have already been used in the in-house data analysis pipeline and by an external website.
To this end, our contributions include: 
\begin{itemize}
    \item Publicly accessible APIs, which address multiple aspects of COVID-19 related disinformation, formulated into seven questions and are in four different languages: Arabic, Bulgarian, Dutch, and English. 
    \item A demo system, which shows individual classified labels, and aggregated statistics over time. 
\end{itemize}

The rest of the paper is organized as follows: Section~\ref{sec:related_works} offers a brief overview of previous work. Section~\ref{sec:disinfo_system} provides a detail of the system. Section~\ref{sec:discussion} discusses possible use cases. Finally, Section~\ref{sec:conclutions} concludes and points to possible directions for future work.

\section{Related Work}
\label{sec:related_works}

There has been a large body of research on fighting the COVID-19 infodemic. Relevant notable work includes studying credibility \cite{cinelli2020covid19,pulido2020covid,zhou2020repository}, racial prejudices and fear \cite{Medford2020.04.03.20052936,vidgen2020detecting},
situational information, e.g.,~caution and advice \cite{li2020characterizing}, as well as on detecting mentions and stance with respect to known misconceptions \cite{hossain-etal-2020-covidlies}. 

\citet{li2020characterizing} analyzed the Sina Weibo microblogging platform to study different situational information types, e.g.,~``caution and advice'', ``donations of money, goods, or services'', ``help seeking'', ``counter-rumor'', etc. \citet{cinelli2020covid19} reported media bias and rumor amplification patterns for COVID-19 using five different social media platforms. \citet{Medford2020.04.03.20052936} analyzed COVID-19 related tweets to understand different content types such as emotional, racially prejudiced, xenophobic or content that causes fear. Other recent work includes identifying low-credibility information using data from social media \citep{yang2020prevalence}, detecting prejudice \citep{vidgen2020detecting}, finding challenges related to data, tools, and ethical issues~\citep{ding2020challenges}, analyzing the spread of COVID-19 misinformation in relation to culture, society, and politics \citep{leng2021misinformation}, detecting the spread of misleading information and the credibility of users who propagate it \citep{mourad2020critical}, identifying positive influencers to propagate information \citep{pastorescuredo2020characterizing}, analyzing the users who spread misinformation and the propagation of misinformation \citep{shahi2021exploratory}, analyzing psychometric aspects in relation to the COVID-19 infodemic \citep{aggrawal2021psychometric}, developing a multilingual COVID-19 Instagram dataset \citep{zarei2020instagram}, and detecting disinformation campaigns \citep{vargas2020detection}. \citet{saakyan-etal-2021-covid} developed a fact extraction and verification dataset consisting of 4,086 claims about the COVID-19 pandemic, collected from Reddit. 

The above research has focused on addressing one or more aspects of the infodemic (e.g., factuality). The work by \citet{song2021classification} addresses several aspects of COVID-19 related disinformation, where they collected false and misleading claims about COVID-19 from IFCN Poynter and annotated them as, \textit{public authority, community spread and impact, medical advice, self-treatments, and virus effects, prominent actors, conspiracies, virus transmission, virus origins and properties, public reaction, and vaccines, medical treatments,} and tests. The authors also developed models and APIs for public access.\footnote{\url{https://github.com/GateNLP/CANTM}}

There has also been work with promising systems, which facilitate users for further analysis of COVID-19 related information such as the work of \citet{zhu-etal-2021-dashboard}, Poynter,\footnote{\url{https://www.poynter.org/ifcn-covid-19-misinformation/}} fake news debunker by InVID \& WeVerify,\footnote{\url{https://www.invid-project.eu/tools-and-services/invid-verification-plugin/}} and google fact-check explorer.\footnote{\url{https://toolbox.google.com/factcheck/explorer}}

Unlike these efforts, our system can support multiple aspects, and we were inspired from prior work where a multi-question annotation schema has been proposed, each of which with 2--10 labels \cite{alam2020call2arms}, and also developed a dataset covering four languages about COVID-19 tweets following the same annotation schema~\cite{alam2020fighting}.

\section{System Overview}
\label{sec:disinfo_system}

Our goal is to provide ready-to-use API services that can be easily integrated into existing systems. We also offer a web demo service to test the system online. Figure \ref{fig:api_services} shows the architecture of the COVID-19 infodemic API services and demo pipeline, which consists of several components, such as classification models with seven questions discussed in section \ref{sssec:data}, API services module, and a demo module. API endpoints serve API post/get requests. The demo pipeline module consists of several parts: \textit{data processing module}, which collects, filter, classify and store the tweets in the Elasticsearch database, the \textit{visualization module} enables to take input (i.e., keywords, dates) (see Section \ref{ssec:online_demo}) from the user and query the Elasticsearch database to visualize the aggregated output. We describe each of them in detail below.

\subsection{Classification Models}
\label{ssec:classification}

\subsubsection{Data}
\label{sssec:data}

For this study, we used the \textit{COVID-19 Disinfo dataset} discussed in~\cite{alam2020fighting}. The dataset consists of annotated tweets in four languages (Arabic, Bulgarian, Dutch, and English), which were originally collected from Twitter using a set of \textit{COVID-19} and \textit{COVID-19-vaccine} related keywords, and different time frames: from January 2020 till March 2021. The annotation of the dataset consists of seven questions as described below: 
\begin{enumerate}
	\item \textbf{Verifiable factual claim:} Does the tweet contain a verifiable factual claim? It consists of two labels: {\em(i)} yes, {\em(ii)} no.
	
	\item \textbf{False information:} To what extent does the tweet appear to contain false information? It has five labels: {\em(i)} NO, definitely contains no false information, {\em(ii)} NO, probably contains no false information, {\em(iii)} Not sure, {\em(iv)} YES, probably contains false information, {\em(v)} YES, definitely contains false information.
	
	\item \textbf{Interest to the general public:} Will the tweet's claim have an impact on or be of interest to the general public? It also has five labels: {\em(i)} NO, definitely not of interest {\em(ii)} NO, probably not of interest, {\em(iii)} Not sure, {\em(iv)} YES, probably of interest, {\em(v)} YES, definitely of interest
	
	\item \textbf{Harmfulness:} To what extent does the tweet appear to be harmful to society, person(s), company(s) or product(s)? Five labels for this question include: {\em(i)} NO, definitely not harmful, {\em(ii)} NO, probably not harmful, {\em(iii)} Not sure, {\em(iv)} YES, probably harmful, {\em(v)} YES, definitely harmful.
	
	\item \textbf{Need for Verification:} Do you think that a professional fact-checker should verify the claim in the tweet? It also consists of five labels: {\em(i)} NO, no need to check, {\em(ii)} NO, too trivial to check, {\em(iii)} YES, not urgent,  {\em(iv)} YES, very urgent, {\em(v)} Not sure.
	
	\item \textbf{Harmful to society:} Is the tweet harmful to society and why? It consists of eight labels:
	{\em(i)} NO, not harmful, {\em(ii)} NO, joke or sarcasm,  {\em(iii)} Not sure, {\em(iv)} YES, panic, {\em(v)} YES, xenophobic, racist, prejudices, or hate-speech, {\em(vi)} YES, bad cure, {\em(vii)} YES, rumor, or conspiracy, {\em(viii)} YES, other
	\item \textbf{Requires attention:} Do you think that this tweet should get the attention of policy makers of government entities? It consists of eight labels: {\em(i)} \emph{No, not interesting}, {\em(ii)} \emph{not sure}, {\em(iii)} \emph{Yes, asks question}, {\em(iv)} \emph{Yes, blame authorities}, {\em(v)} \emph{Yes, calls for action}, {\em(vi)} \emph{Yes, classified as in harmful task}, {\em(vii)} \emph{Yes, contains advice}, {\em(viii)} \emph{Yes, discusses action taken}, {\em(ix)} \emph{Yes, discusses cure}, and {\em(x)} \emph{Yes, other}.
\end{enumerate}

The annotated dataset consists of a total of 4,966, 3,697, 2,665, and 4,542 tweets for Arabic, Bulgarian, Dutch, and English, respectively. Table~\ref{tab:class_label_distribution} shows the distribution of the class labels for all languages.

\begin{table*}[!tbh]
\centering
\setlength{\tabcolsep}{2.0pt}  
\scalebox{0.90}{
\begin{tabular}{lrrrrr}
\toprule
\multicolumn{1}{c}{\textbf{Question}} & \multicolumn{1}{c}{\textbf{Cl.}} & \multicolumn{1}{c}{\textbf{Arabic}} & \multicolumn{1}{c}{\textbf{Bulgarian}} & \multicolumn{1}{c}{\textbf{Dutch}} & \multicolumn{1}{c}{\textbf{English}} \\ \midrule
Q1: Verifiable factual claim & 2 & 4,966 & 3,697 & 2,665 & 4,542 \\
Q2: False information & 5 & 3,439 & 2,567 & 1,253 & 2,891 \\
Q3: Interest to the general public & 5 & 3,439 & 2,567 & 1,253 & 2,891 \\
Q4: Harmfulness & 5 & 3,439 & 2,567 & 1,253 & 2,891 \\
Q5: Need for verification & 5 & 3,439 & 2,567 & 1,247 & 2,891 \\
Q6: Harmful to society & 8 & 4,966 & 3,697 & 2,665 & 4,542 \\
Q7: Requires attention & 10 & 4,966 & 3,697 & 2,665 & 4,542 \\
\bottomrule
\end{tabular}
}
\caption{Number of annotated tweets per language and per question.}
\label{tab:class_label_distribution}
\end{table*}

\begin{table*}[!tbh]
\centering
\setlength{\tabcolsep}{2.0pt}  
\scalebox{0.90}{
\begin{tabular}{@{}lrrrrrrrrr@{}}
\toprule
\multicolumn{1}{c}{\textbf{}} & \multicolumn{1}{c}{\textbf{}} & \multicolumn{2}{c}{\textbf{Arabic}} & \multicolumn{2}{c}{\textbf{Bulgarian}} & \multicolumn{2}{c}{\textbf{Dutch}} & \multicolumn{2}{c}{\textbf{English}} \\ \midrule
\multicolumn{1}{c}{\textbf{Question}} & \multicolumn{1}{c}{\textbf{Cl.}} & \multicolumn{1}{c}{\textbf{Maj}} & \multicolumn{1}{c}{\textbf{SVM}} & \multicolumn{1}{c}{\textbf{Maj}} & \multicolumn{1}{c}{\textbf{SVM}} & \multicolumn{1}{c}{\textbf{Maj}} & \multicolumn{1}{c}{\textbf{SVM}} & \multicolumn{1}{c}{\textbf{Maj}} & \multicolumn{1}{c}{\textbf{SVM}} \\ \midrule
\multicolumn{10}{c}{\textbf{Binary}} \\ \midrule
Q1: Verifiable factual claim & 2 & 56.8 & \textbf{80.5} & 58.3 & \textbf{77.2} & 36.5 & \textbf{64.6} & 48.7 & \textbf{68.5} \\
Q2: False information & 2 & 68.3 & \textbf{80.4} & \textbf{95.0} & \textbf{95.0} & 64.9 & \textbf{76.6} & \textbf{91.6} & 91.2 \\
Q3: Interest to the general public & 2 & \textbf{96.3} & \textbf{96.3} & \textbf{96.5} & \textbf{96.5} & 62.3 & \textbf{70.8} & \textbf{96.3} & 96.1 \\
Q4: Harmfulness & 2 & 67.2 & \textbf{84.1} & \textbf{86.8} & \textbf{86.8} & 63.9 & \textbf{72.4} & 66.7 & \textbf{79.3} \\
Q5: Need for verification & 2 & 46.8 & \textbf{60.9} & 70.5 & \textbf{74.9} & 44.4 & \textbf{63.2} & 67.7 & \textbf{77.8} \\
Q6: Harmful to society & 2 & 72.5 & \textbf{84.5} & 83.2 & \textbf{83.5} & 84.7 & \textbf{86.4} & \textbf{86.7} & 85.6 \\
Q7: Requires attention & 2 & 57.7 & \textbf{73.0} & \textbf{80.1} & 79.5 & 65.6 & \textbf{76.3} & 78.3 & \textbf{83.4} \\ \midrule
\multicolumn{10}{c}{\textbf{Multiclass}} \\ \midrule
Q2: False information & 5 & 62.9 & \textbf{80.5} & 77.3 & \textbf{79.0} & 36.5 & \textbf{64.6} & 67.9 & \textbf{68.5} \\
Q3: Interest to the general public & 5 & 44.4 & \textbf{72.1} & 64.2 & \textbf{70.1} & 32.0 & \textbf{41.7} & \textbf{78.9} & 69.6 \\
Q4: Harmfulness & 5 & 28.1 & \textbf{56.7} & 58.8 & \textbf{64.1} & 21.0 & \textbf{49.1} & 19.9 & \textbf{82.8} \\
Q5: Need for verification & 5 & 41.2 & \textbf{46.6} & 36.0 & \textbf{56.0} & 18.4 & \textbf{43.7} & 46.8 & \textbf{48.2} \\
Q6: Harmful to society & 8 & \textbf{68.7} & 49.8 & 76.6 & \textbf{76.9} & \textbf{74.4} & 41.1 & \textbf{84.0} & 57.6 \\
Q7: Requires attention & 10 & 13.8 & \textbf{78.2} & \textbf{80.1} & 79.0 & 65.4 & \textbf{76.3} & 78.1 & \textbf{84.9} \\ \bottomrule
\end{tabular}
	}
\caption{Performance of our models for COVID-19 infodamic. Here, \emph{\# Cl.} shows the number of classes for the corresponding question.}
\label{tab:disinfo_classification_results}
\end{table*}

\subsubsection{Data Preparation} 

To train the models, we generated a stratified split \cite{sechidis2011stratification} of the data into 70\%/10\%/20\% for training/development/testing, respectively. We then preprocess the data, which includes removal of hash-symbols and non-alphanumeric symbols, case folding, URL replacement with a URL tag, and username replacement with a user tag. 

\subsubsection{Models} 
We train SVM classifiers, where we optimized the hyper-parameters using the development sets. Our choice of SVM models for deployment was due to their computational simplicity. \citet{alam2020fighting} described transformer-based models where reported results were comparatively higher than SVM models. We could not deploy such models due to their expensive hardware requirements, which we plan to include in a future deployment. 

The annotation schema and the dataset have been designed in a way that they can easily be converted into binary labels. Hence, we also mapped the fine-grained labels to binary labels. In addition to training the models in multiclass settings, we also trained the models in the binary setting. 

\subsubsection{Results} 
Table~\ref{tab:disinfo_classification_results} shows the performance of the classifiers for all languages: both in binary and in multiclass settings. We report the performance in terms of weighted average, which takes into account class imbalance issue. All results are higher than the majority baseline, except for a few cases where the class label distribution is very skewed (e.g., for Bulgarian, Q7 with binary labels). The results in the binary settings are higher than in multiclass settings, which is a typical scenario for the classification models. Note the multiclass nature of the tasks and the skewed class distribution for Q2 to Q7~\cite{alam2020fighting}, has an impact on the classification performance.  

\subsection{API Services}
The API services provide the functionality to process and classify language-specific tweets by specifying the desired language as shown in Figure \ref{fig:system_api}. The \textit{post} call of the API service takes a tweet text, language (i.e., Arabic, Bulgarian, Dutch, or English), and task (i.e., binary or multiclass) as parameters and returns a \textit{key} with a success message. 
Then the \textit{get} call of the API service takes the \textit{key} and language as parameters and returns the results, which consists of label and probability for each question. In addition, we also provide a label dictionary for each question, which consists of the label and probability for each class. This can facilitate users to further post-process on the labels and the probability scores if needed. The API services are publicly available and can be tested on \url{https://app.swaggerhub.com/apis/yifan2019/Tanbih/0.8.0} under \textit{post/get\_covid19disformation}, as also depicted in Figure \ref{fig:system_api}. We also provide relevant API client scripts publicly to facilitate users.\footnote{https://github.com/firojalam/covid19-infodemic-demo}

\begin{figure*}[!tbh]
	\centering
    \begin{subfigure}[b]{0.42\textwidth}    
        \includegraphics[width=\textwidth]{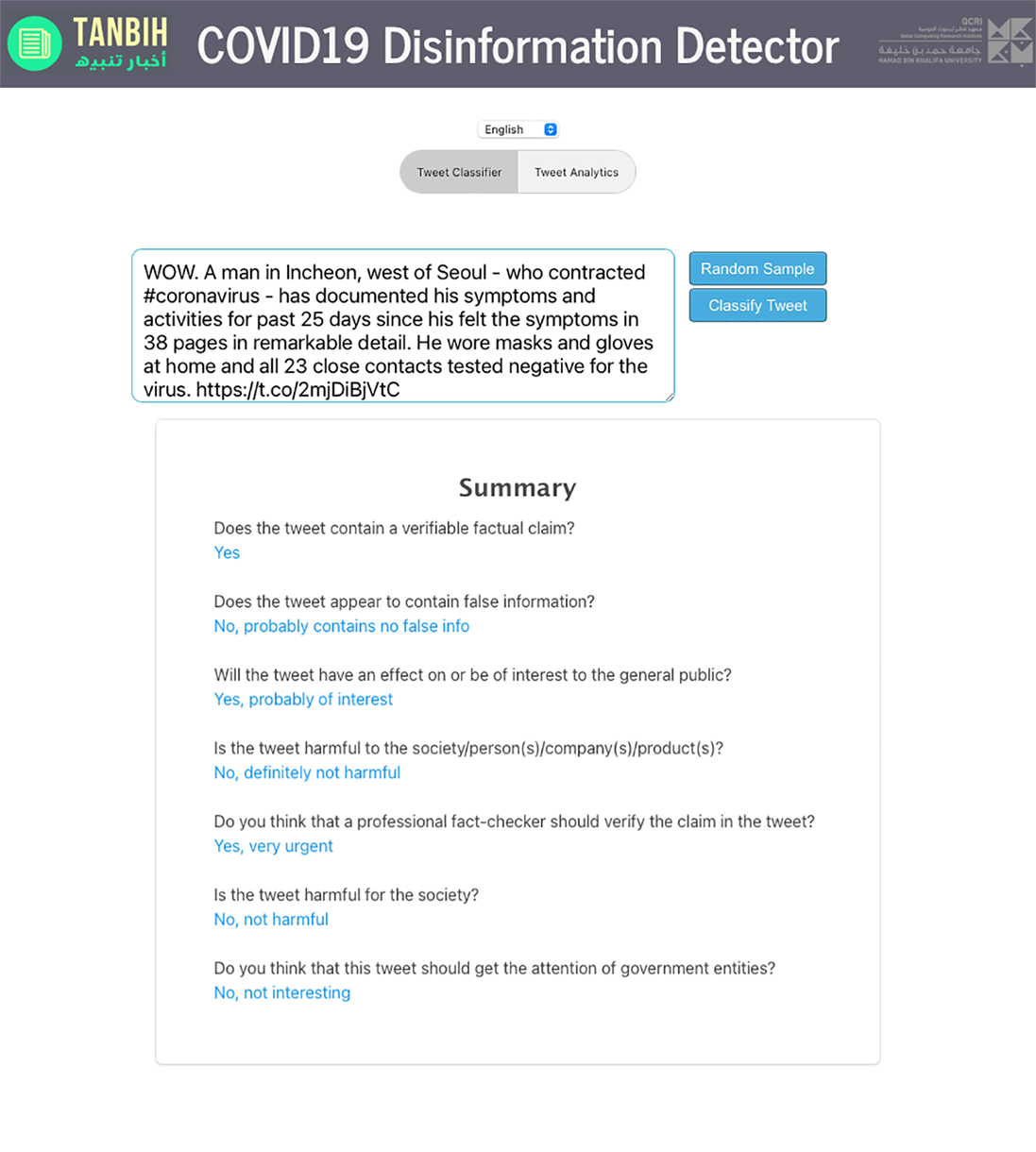}
        \caption{Individual tweet classification.}
        \label{fig:demo_single_tweet}
    \end{subfigure}
    \begin{subfigure}[b]{0.42\textwidth}    
        \includegraphics[width=\textwidth]{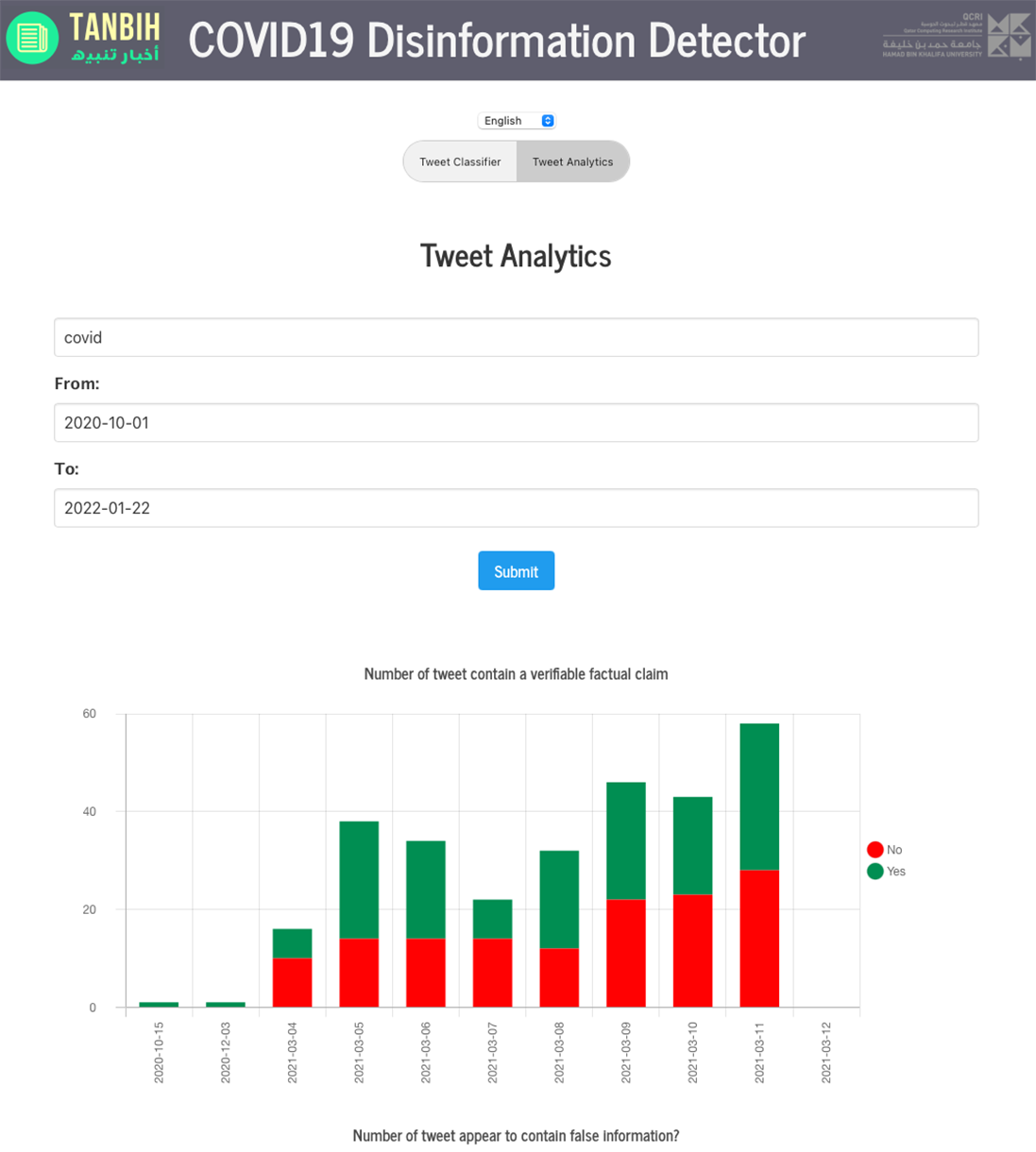}
        \caption{Analytics.}
        \label{fig:demo_aggr_stat}
    \end{subfigure} 
    \caption{Cropped version of the online demo interface.}
    \label{fig:system_demo}
\end{figure*}

\subsection{Online Demo}
\label{ssec:online_demo}

The demo interface has two parts: tweet classifier and tweet analytics as shown in Figure \ref{fig:system_demo}. An option to select a language enables to direct the demo to process language-specific content. 

In the \textit{Tweet Classifier} part, a text area allows the user to input text, and the \textit{classify tweet} option allows to show classified results as a summary and individual plots to visualize the class-specific probabilities.

The purpose of \textit{Tweet Analytics} part, as shown in Figure \ref{fig:demo_aggr_stat}, is to visualize aggregated statistics over time to understand whether certain phenomena increasing or decreasing. For example, whether tweets with jokes and/or rumors increase or decrease over time? \citet{nakov-etal-2021-second} highlighted that in two different time periods (February’2020 – August’2020 \textit{vs.} November’2020 – January’2021), there were many jokes, rarely rumors, and many tweets with factually true claims, which is analyzed based on tweets posted in Qatar.

For the \textit{Tweet Analytics} demo, we collected tweets since February 2020 using COVID-19 related keywords. Our data collection consists of tweets in four different languages as mentioned earlier. The \textit{dataset processing module}, as shown in Figure \ref{fig:api_services}, filter tweets with different criteria such as {\em(i)} tokenize tweets and check if number of token is less than five, {\em(ii)} detect language using language detection tool.\footnote{\url{https://github.com/pemistahl/lingua}},\footnote{Note that Twitter language tag is not always reliable as also discussed in \cite{alam2020standardizing}} Moreover, we also extract full text instead of truncated text that appears in the text field of the Twitter JSON object. 
The \textit{dataset processing module} then uses API services to classify tweets and to store them in the Elasticsearch database, which provides text search functionality in addition to other search criteria and its capability to deal with a large dataset. Finally, \textit{the visualization module} deals with user query, i.e., search string, dates to visualize the aggregated statistics, as shown in Figure \ref{fig:demo_aggr_stat}. The \textit{analytics part} visualize the day-wise statistics for each question.

\section{Discussion}
\label{sec:discussion}

The API services can be used in different application scenarios. For example, analyzing tweets with trending topics, which might be specific to any event. Finding content that can spread panic or rumor, or blames authorities are of most interest to policymakers as it can help them to make actionable decision. Another use case could be to detect tweets as early as possible that can spread panic, rumors, or false claims, so that they can be debunked.

\section{Conclusion and Future Work}
\label{sec:conclutions}

We have presented a system to fight the global COVID-19 infodemic. We provide API services, which support four different languages, for the community and non-technical end-users. The API services are freely available and support both binary and fine-grained classification. In addition, we provide an online demo to test the system online. 

In the future, we plan to include transformer-based models as part of the API services and improve the analytics part to support country-specific information. Such functionality would be more useful to a large number of user types: practitioners, professional fact-checker, journalists, social media platforms, and policymakers.

\section*{Ethics Statement}
We collected the dataset using the Twitter API\footnote{\url{http://developer.twitter.com/en/docs}} with keywords that only use terms related to COVID-19, without other biases. We followed the terms of use outlined by Twitter.\footnote{\url{http://developer.twitter.com/en/developer-terms/agreement-and-policy}} Specifically, we only downloaded public tweets. 

The API services can be used to analyse social media content, which could be of interest to practitioners, professional fact-checker, journalists, social media platforms, and policymakers. The output of the services can be used to alleviate the burden for social media moderators, but human supervision would be required for more intricate cases and in order to ensure that the system does not cause harm.

Our API services can help fight the COVID-19 infodemic, and they could support analysis and decision making for the public good. However, the output of the  API services can also be misused by malicious actors. Therefore, we ask the potential users to be aware of potential misuse.


\bibliography{bib/main}

\begin{thebibliography}{24}
\expandafter\ifx\csname natexlab\endcsname\relax\def\natexlab#1{#1}\fi

\bibitem[{Aggrawal et~al.(2021)Aggrawal, Jolly, Gulati, Sethi, Kumaraguru, and
  Sethi}]{aggrawal2021psychometric}
Palash Aggrawal, Baani Leen~Kaur Jolly, Amogh Gulati, Amarjit Sethi,
  Ponnurangam Kumaraguru, and Tavpritesh Sethi. 2021.
\newblock Psychometric analysis and coupling of emotions between state
  bulletins and twitter in india during covid-19 infodemic.
\newblock \emph{Frontiers in Communication}, page 158.

\bibitem[{Alam et~al.(2021{\natexlab{a}})Alam, Dalvi, Shaar, Durrani, Mubarak,
  Nikolov, {Da San Martino}, Abdelali, Sajjad, Darwish, and
  Nakov}]{alam2020fighting}
Firoj Alam, Fahim Dalvi, Shaden Shaar, Nadir Durrani, Hamdy Mubarak, Alex
  Nikolov, Giovanni {Da San Martino}, Ahmed Abdelali, Hassan Sajjad, Kareem
  Darwish, and Preslav Nakov. 2021{\natexlab{a}}.
\newblock \href {https://ojs.aaai.org/index.php/ICWSM/article/view/18114}
  {Fighting the {COVID}-19 infodemic in social media: A holistic perspective
  and a call to arms}.
\newblock In \emph{Proceedings of the International {AAAI} Conference on Web
  and Social Media}, ICWSM~'21, pages 913--922.

\bibitem[{Alam et~al.(2021{\natexlab{b}})Alam, Dalvi, Shaar, Durrani, Mubarak,
  Nikolov, {Da San Martino}, Abdelali, Sajjad, Darwish, and
  Nakov}]{alam2020call2arms}
Firoj Alam, Fahim Dalvi, Shaden Shaar, Nadir Durrani, Hamdy Mubarak, Alex
  Nikolov, Giovanni {Da San Martino}, Ahmed Abdelali, Hassan Sajjad, Kareem
  Darwish, and Preslav Nakov. 2021{\natexlab{b}}.
\newblock Fighting the {COVID}-19 infodemic in social media: A holistic
  perspective and a call to arms.
\newblock In \emph{Proceedings of the International {AAAI} Conference on Web
  and Social Media}, ICWSM~'21, pages 913--922.

\bibitem[{Alam et~al.(2021{\natexlab{c}})Alam, Sajjad, Imran, and
  Ofli}]{alam2020standardizing}
Firoj Alam, Hassan Sajjad, Muhammad Imran, and Ferda Ofli. 2021{\natexlab{c}}.
\newblock \href {https://ojs.aaai.org/index.php/ICWSM/article/view/18115}
  {{CrisisBench}: Benchmarking crisis-related social media datasets for
  humanitarian information processing}.
\newblock In \emph{Proceedings of the International AAAI Conference on Web and
  Social Media}, ICWSM~'21, pages 923--932.

\bibitem[{Cinelli et~al.(2020)Cinelli, Quattrociocchi, Galeazzi, Valensise,
  Brugnoli, Schmidt, Zola, Zollo, and Scala}]{cinelli2020covid19}
Matteo Cinelli, Walter Quattrociocchi, Alessandro Galeazzi, Carlo~Michele
  Valensise, Emanuele Brugnoli, Ana~Lucia Schmidt, Paola Zola, Fabiana Zollo,
  and Antonio Scala. 2020.
\newblock \href {https://www.nature.com/articles/s41598-020-73510-5} {The
  {COVID-19} social media infodemic}.
\newblock \emph{Scientific Reports}, 10(1):1--10.

\bibitem[{Ding et~al.(2020)Ding, Shu, Li, Bhattacharjee, and
  Liu}]{ding2020challenges}
Kaize Ding, Kai Shu, Yichuan Li, Amrita Bhattacharjee, and Huan Liu. 2020.
\newblock Challenges in combating covid-19 infodemic -- data, tools, and
  ethics.
\newblock In \emph{Proceedings of the CIKM 2020 Workshops}, CIKM~'20.

\bibitem[{Hossain et~al.(2020)Hossain, Logan~IV, Ugarte, Matsubara, Young, and
  Singh}]{hossain-etal-2020-covidlies}
Tamanna Hossain, Robert~L. Logan~IV, Arjuna Ugarte, Yoshitomo Matsubara, Sean
  Young, and Sameer Singh. 2020.
\newblock \href {https://www.aclweb.org/anthology/2020.nlpcovid19-2.11}
  {{COVIDL}ies: Detecting {COVID}-19 misinformation on social media}.
\newblock In \emph{Proceedings of the 1st Workshop on {NLP} for {COVID}-19
  (Part 2) at {EMNLP} 2020}, Online. Association for Computational Linguistics.

\bibitem[{Leng et~al.(2021)Leng, Zhai, Sun, Wu, Selzer, Strover, Zhang, Chen,
  and Ding}]{leng2021misinformation}
Yan Leng, Yujia Zhai, Shaojing Sun, Yifei Wu, Jordan Selzer, Sharon Strover,
  Hezhao Zhang, Anfan Chen, and Ying Ding. 2021.
\newblock Misinformation during the covid-19 outbreak in china: Cultural,
  social and political entanglements.
\newblock \emph{IEEE Transactions on Big Data}, 7(1):69--80.

\bibitem[{Li et~al.(2020)Li, Zhang, Wang, Zhang, Wang, Gao, Duan, Tsoi, and
  Wang}]{li2020characterizing}
Lifang Li, Qingpeng Zhang, Xiao Wang, Jun Zhang, Tao Wang, Tian-Lu Gao, Wei
  Duan, Kelvin Kam-fai Tsoi, and Fei-Yue Wang. 2020.
\newblock \href {https://doi.org/10.1109/TCSS.2020.2980007} {Characterizing the
  propagation of situational information in social media during {COVID-19}
  epidemic: A case study on {Weibo}}.
\newblock \emph{IEEE Transactions on Computational Social Systems},
  7(2):556--562.

\bibitem[{Medford et~al.(2020)Medford, Saleh, Sumarsono, Perl, and
  Lehmann}]{Medford2020.04.03.20052936}
Richard~J Medford, Sameh~N Saleh, Andrew Sumarsono, Trish~M Perl, and
  Christoph~U Lehmann. 2020.
\newblock \href {https://doi.org/10.1093/ofid/ofaa258} {{An ``Infodemic'':
  Leveraging High-Volume {T}witter Data to Understand Early Public Sentiment
  for the Coronavirus Disease 2019 Outbreak}}.
\newblock \emph{Open Forum Infectious Diseases}, 7(7).
\newblock Ofaa258.

\bibitem[{Mourad et~al.(2020)Mourad, Srour, Harmanai, Jenainati, and
  Arafeh}]{mourad2020critical}
Azzam Mourad, Ali Srour, Haidar Harmanai, Cathia Jenainati, and Mohamad Arafeh.
  2020.
\newblock Critical impact of social networks infodemic on defeating coronavirus
  covid-19 pandemic: Twitter-based study and research directions.
\newblock \emph{IEEE Transactions on Network and Service Management},
  17(4):2145--2155.

\bibitem[{Nakov et~al.(2021)Nakov, Alam, Shaar, Da~San~Martino, and
  Zhang}]{nakov-etal-2021-second}
Preslav Nakov, Firoj Alam, Shaden Shaar, Giovanni Da~San~Martino, and Yifan
  Zhang. 2021.
\newblock A second pandemic? analysis of fake news about {COVID}-19 vaccines in
  {Q}atar.
\newblock In \emph{Proc. of Conference on Recent Advances in Natural Language
  Processing}, pages 1010--1021.

\bibitem[{Pastor-Escuredo and
  Tarazona(2020)}]{pastorescuredo2020characterizing}
David Pastor-Escuredo and Carlota Tarazona. 2020.
\newblock \href {http://arxiv.org/abs/2005.07266} {Characterizing information
  leaders in twitter during covid-19 crisis}.

\bibitem[{Pulido et~al.(2020)Pulido, Villarejo-Carballido, Redondo-Sama, and
  Gómez}]{pulido2020covid}
Cristina~M Pulido, Beatriz Villarejo-Carballido, Gisela Redondo-Sama, and Aitor
  Gómez. 2020.
\newblock \href {https://doi.org/10.1177/0268580920914755} {{COVID-19}
  infodemic: More retweets for science-based information on coronavirus than
  for false information}.
\newblock \emph{International Sociology}, 35(4):377--392.

\bibitem[{Saakyan et~al.(2021)Saakyan, Chakrabarty, and
  Muresan}]{saakyan-etal-2021-covid}
Arkadiy Saakyan, Tuhin Chakrabarty, and Smaranda Muresan. 2021.
\newblock \href {https://doi.org/10.18653/v1/2021.acl-long.165} {{COVID}-fact:
  Fact extraction and verification of real-world claims on {COVID}-19
  pandemic}.
\newblock In \emph{Proceedings of the 59th Annual Meeting of the Association
  for Computational Linguistics and the 11th International Joint Conference on
  Natural Language Processing (Volume 1: Long Papers)}, pages 2116--2129,
  Online. Association for Computational Linguistics.

\bibitem[{Sechidis et~al.(2011)Sechidis, Tsoumakas, and
  Vlahavas}]{sechidis2011stratification}
Konstantinos Sechidis, Grigorios Tsoumakas, and Ioannis Vlahavas. 2011.
\newblock On the stratification of multi-label data.
\newblock In \emph{Machine Learning and Knowledge Discovery in Databases},
  ECML-PKDD~'11, pages 145--158, Berlin, Heidelberg. Springer Berlin
  Heidelberg.

\bibitem[{Shahi et~al.(2021)Shahi, Dirkson, and
  Majchrzak}]{shahi2021exploratory}
Gautam~Kishore Shahi, Anne Dirkson, and Tim~A Majchrzak. 2021.
\newblock An exploratory study of covid-19 misinformation on twitter.
\newblock \emph{Online social networks and media}, 22:100104.

\bibitem[{Song et~al.(2021)Song, Petrak, Jiang, Singh, Maynard, and
  Bontcheva}]{song2021classification}
Xingyi Song, Johann Petrak, Ye~Jiang, Iknoor Singh, Diana Maynard, and Kalina
  Bontcheva. 2021.
\newblock Classification aware neural topic model for covid-19 disinformation
  categorisation.
\newblock \emph{PloS one}, 16(2):e0247086.

\bibitem[{Vargas et~al.(2020)Vargas, Emami, and Traynor}]{vargas2020detection}
Luis Vargas, Patrick Emami, and Patrick Traynor. 2020.
\newblock On the detection of disinformation campaign activity with network
  analysis.
\newblock In \emph{Proceedings of the 2020 ACM SIGSAC Conference on Cloud
  Computing Security Workshop}, pages 133--146.

\bibitem[{Vidgen et~al.(2020)Vidgen, Hale, Guest, Margetts, Broniatowski,
  Waseem, Botelho, Hall, and Tromble}]{vidgen2020detecting}
Bertie Vidgen, Scott Hale, Ella Guest, Helen Margetts, David Broniatowski,
  Zeerak Waseem, Austin Botelho, Matthew Hall, and Rebekah Tromble. 2020.
\newblock \href {https://doi.org/10.18653/v1/2020.alw-1.19} {Detecting {E}ast
  {A}sian prejudice on social media}.
\newblock In \emph{Proceedings of the Fourth Workshop on Online Abuse and
  Harms}, ALW~'20, pages 162--172, Online. Association for Computational
  Linguistics.

\bibitem[{Yang et~al.(2020)Yang, Torres-Lugo, and Menczer}]{yang2020prevalence}
Kai-Cheng Yang, Christopher Torres-Lugo, and Filippo Menczer. 2020.
\newblock Prevalence of low-credibility information on twitter during the
  covid-19 outbreak.
\newblock In \emph{Proceedings of the Ninth International Conference on Web and
  Social Media}, ICWSM~'20. {AAAI} Press.

\bibitem[{Zarei et~al.(2020)Zarei, Farahbakhsh, Crespi, and
  Tyson}]{zarei2020instagram}
Koosha Zarei, Reza Farahbakhsh, Noel Crespi, and Gareth Tyson. 2020.
\newblock \href {http://arxiv.org/abs/2004.12226} {A first instagram dataset on
  covid-19}.

\bibitem[{Zhou et~al.(2020)Zhou, Mulay, Ferrara, and
  Zafarani}]{zhou2020repository}
Xinyi Zhou, Apurva Mulay, Emilio Ferrara, and Reza Zafarani. 2020.
\newblock \href {https://arxiv.org/abs/2006.05557} {{ReCOVery}: A multimodal
  repository for {COVID-19} news credibility research}.
\newblock In \emph{Proceedings of the 29th ACM International Conference on
  Information \& Knowledge Management}, CIKM~'20, pages 3205--3212. Association
  for Computing Machinery.

\bibitem[{Zhu et~al.(2021)Zhu, Meng, Caraballo, Jaradat, Shi, Zhang, Akrami,
  Liao, Arslan, Jimenez, Saeef, Pathak, and Li}]{zhu-etal-2021-dashboard}
Zhengyuan Zhu, Kevin Meng, Josue Caraballo, Israa Jaradat, Xiao Shi, Zeyu
  Zhang, Farahnaz Akrami, Haojin Liao, Fatma Arslan, Damian Jimenez,
  Mohanmmed~Samiul Saeef, Paras Pathak, and Chengkai Li. 2021.
\newblock \href {https://aclanthology.org/2021.eacl-demos.12} {A dashboard for
  mitigating the {COVID}-19 misinfodemic}.
\newblock In \emph{Proceedings of the 16th Conference of the European Chapter
  of the Association for Computational Linguistics: System Demonstrations},
  pages 99--105, Online. Association for Computational Linguistics.

\end{thebibliography}
\bibliographystyle{acl_natbib}

\end{document}